\documentclass[sigconf]{acmart}

\AtBeginDocument{%
  }

\copyrightyear{2024}
\acmYear{2024}
\setcopyright{rightsretained}
\acmConference[MM '24]{Proceedings of the 32nd ACM International Conference on Multimedia}{October 28-November 1, 2024}{Melbourne, VIC, Australia}
\acmBooktitle{Proceedings of the 32nd ACM International Conference on Multimedia (MM '24), October 28-November 1, 2024, Melbourne, VIC, Australia}
\acmDOI{10.1145/3664647.3680653}
\acmISBN{979-8-4007-0686-8/24/10}

\makeatletter
\gdef\@copyrightpermission{
  \begin{minipage}{0.3\columnwidth}
   \href{https://creativecommons.org/licenses/by/4.0/}{\includegraphics[width=0.90\textwidth]{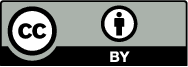}}
  \end{minipage}\hfill
  \begin{minipage}{0.7\columnwidth}
   \href{https://creativecommons.org/licenses/by/4.0/}{This work is licensed under a Creative Commons Attribution International 4.0 License.}
  \end{minipage}
  \vspace{5pt}
}
\makeatother

\usepackage{array}
\usepackage{amsfonts}
\usepackage{amsmath}
\usepackage{booktabs}
\usepackage{multirow}
\usepackage{float}
\usepackage{algorithm}
\usepackage{algorithmic}
\usepackage{hyperref}
\usepackage[capitalize]{cleveref}
\usepackage{enumerate}
\usepackage{enumitem}
\usepackage{makecell}
\usepackage{pbalance}
\usepackage{lineno}

\begin{document}

\title[Robust Multimodal Sentiment Analysis with Distribution-Based Feature Recovery and Fusion]{Robust Multimodal Sentiment Analysis of Image-Text Pairs by Distribution-Based Feature Recovery and Fusion}

\author{Daiqing Wu}
\affiliation{%
  \institution{Institute of Information Engineering, Chinese Academy of Sciences}
  \institution{School of Cyber Security, University of Chinese Academy of Sciences}
  \city{Beijing}
  \country{China}
  }
\email{wudaiqing@iie.ac.cn}

\author{Dongbao Yang}
\authornote{Dongbao Yang and Can Ma are the corresponding authors.}
\affiliation{%
  \institution{Institute of Information Engineering, Chinese Academy of Sciences}
  \city{Beijing}
  \country{China}
  }
\email{yangdongbao@iie.ac.cn}

\author{Yu Zhou}
\affiliation{%
  \institution{TMCC, College of Computer Science, Nankai University}
  \city{Tianjin}
  \country{China}
  }
\email{yzhou@nankai.edu.cn}

\author{Can Ma}
\authornotemark[1]
\affiliation{%
  \institution{Institute of Information Engineering, Chinese Academy of Sciences}
  \city{Beijing}
  \country{China}
  }
\email{macan@iie.ac.cn}

\renewcommand{\shortauthors}{Daiqing Wu, Dongbao Yang, Yu Zhou, \& Can Ma}

\begin{abstract}
As posts on social media increase rapidly, analyzing the sentiments embedded in image-text pairs has become a popular research topic in recent years. Although existing works achieve impressive accomplishments in simultaneously harnessing image and text information, they lack the considerations of possible low-quality and missing modalities. In real-world applications, these issues might frequently occur, leading to urgent needs for models capable of predicting sentiment robustly. Therefore, we propose a \underline{D}istribution-based feature \underline{R}ecovery and \underline{F}usion (DRF) method for robust multimodal sentiment analysis of image-text pairs. Specifically, we maintain a feature queue for each modality to approximate their feature distributions, through which we can simultaneously handle low-quality and missing modalities in a unified framework. For low-quality modalities, we reduce their contributions to the fusion by quantitatively estimating modality qualities based on the distributions. For missing modalities, we build inter-modal mapping relationships supervised by samples and distributions, thereby recovering the missing modalities from available ones. In experiments, two disruption strategies that corrupt and discard some modalities in samples are adopted to mimic the low-quality and missing modalities in various real-world scenarios. Through comprehensive experiments on three publicly available image-text datasets, we demonstrate the universal improvements of DRF compared to SOTA methods under both two strategies, validating its effectiveness in robust multimodal sentiment analysis.
\end{abstract}



\maketitle

\begin{figure}[htbp]
      \centering
      \includegraphics[width=1\linewidth]{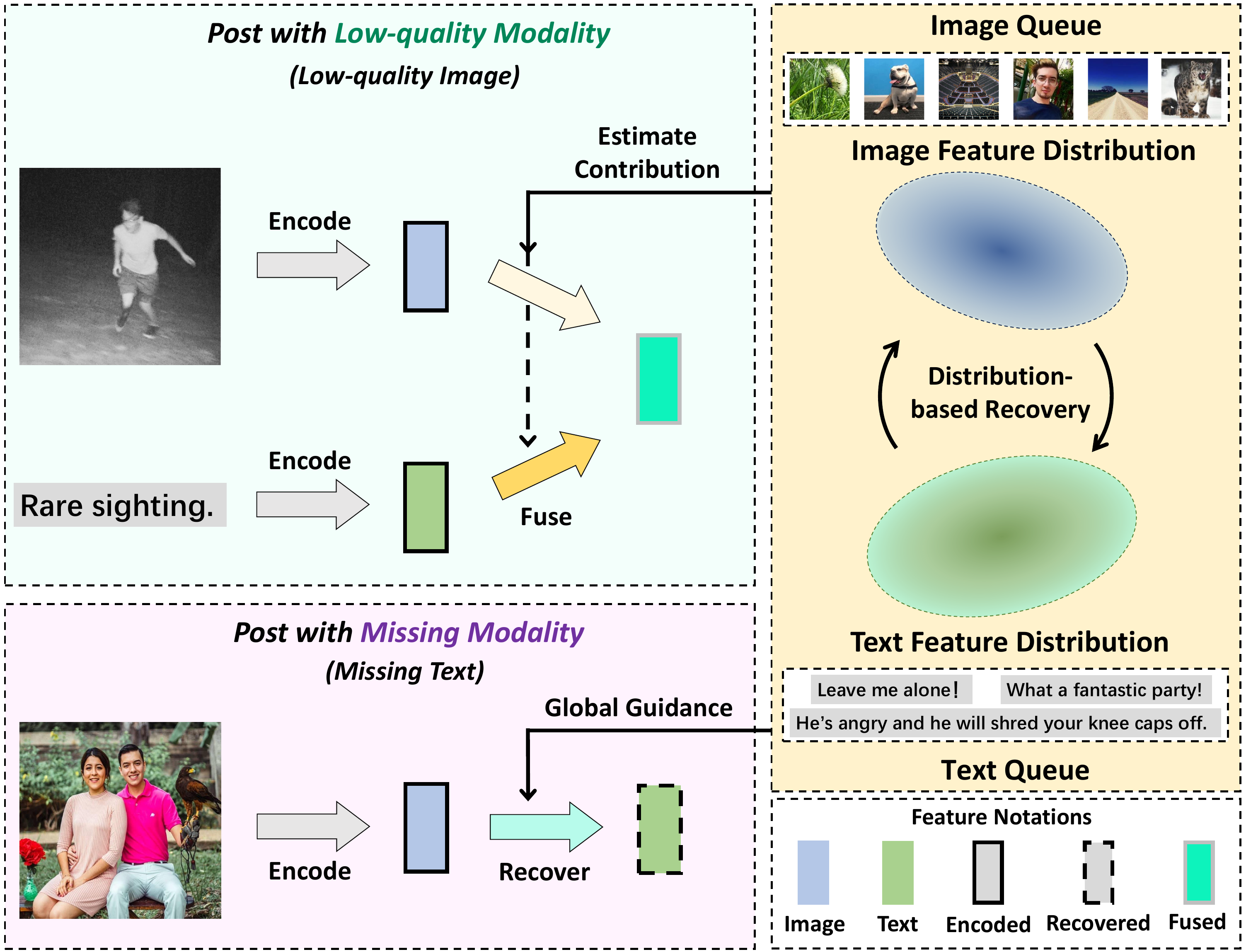}
      \vspace{-15pt}
      \caption{Brief illustration of DRF. We maintain two feature queues to approximate the feature distributions of images and texts. The distributions can estimate the contribution of each modality for fusion and provide global guidance for modality recovery, facilitating the robustness of the model to both low-quality and missing modalities. }
      \label{brief illustration}
      \vspace{-5pt}
\end{figure}


\section{Introduction}
With the rapid growth of smartphones, people are getting used to sharing their experiences by posting on social media. In most cases, posts contain information from various modalities. As a result, multimodal sentiment analysis (MSA) that aims to understand the sentiments expressed by users in multimodal content has become a popular research topic. Due to its wide applications in social media analysis \cite{acl2019hfm}, recommendation system \cite{mm2019dialog}, human-computer interaction \cite{aaai2021hci}, and more \cite{if2021survey, pami2022analysis}, it attracts substantial attention from both academic and industrial communities \cite{zhang2018deep, yue2019survey}.

Image-text pairs are a typical form of posts, and analyzing their overall sentiments is an important subfield in MSA. In existing works, the majority seeks to fuse multimodal information by elaborate fusion strategies, such as concatenations \cite{cikm2017multisentibank} and attentional mechanisms \cite{isi2017hsan, sigir2018comem, tmm2021mvan, naacl2022clmlf}. The others attempt to address task-specific challenges, like the ignorance of global co-occurring characteristics \cite{acl2021mgnns}, modality heterogeneity \cite{acl2023mvcn}, and data dependency \cite{mm2022upmpf, mm2023multipoint}. They achieve impressive progress in fully exploiting information from both visual and textual modalities to model the overall sentiments. However, in real-world applications, the images and texts of posts may be corrupted or missing, leading to frequent occurrences of low-quality and missing modalities. For instance, images are probably pixelated or unavailable due to Not-Safe-For-Work issues and privacy concerns \cite{nips2023incomplete}, and texts perhaps suffer from information loss or are unrecognizable due to rare languages and unaligned encoding formats between platforms. These scenarios result in severe performance degenerations of current works, underscoring the necessity of robust MSA methods.

Handling low-quality or missing modalities has been well-studied in related multimodal fields \cite{mm2021tbfr, emnlp2022mmalign, fcy2022icme, shw2023ijcai, tmm2024nibat, taffc2024emt}. In trusted multi-view classification \cite{iclr2021tmvc, pami2022tmvc}, researchers assign different weights for each view by estimating its uncertainty to produce reliable predictions with potential low-quality views. In incomplete multimodal learning \cite{ijcai2018semi-incomplete, aaai2019cycle-translate, pami2023gcnet}, researchers recover unavailable modalities from the observed ones \cite{nips2023incomplete} to enable consistent encoding of samples with arbitrary missing modalities \cite{nips2019cpmnet}. Despite their success, applying them to handle both issues of low-quality and missing modalities in MSA of image-text pairs would encounter two main challenges. Firstly, the two issues are tackled separately, with unaligned models designed based on distinct strategies, which introduces extra difficulties and alignment burdens for direct combination. Secondly, the user-generated nature of posts from social media results in frequent mismatches between images and texts \cite{acl2021mgnns, wdq2024arxiv}. This characteristic conflicts with the common assumption in studies on videos or medical images \cite{aaai2019cycle-translate, pami2023gcnet, nips2023incomplete}, that the information of modalities from the same sample is closely related, impeding the application of these methods.


To fill these gaps, we propose a method called \underline{D}istribution-based feature \underline{R}ecovery and \underline{F}usion (\textbf{DRF}), as shown in \cref{brief illustration}. We maintain feature queues for images and texts to approximate their respective feature distributions, which enable the model to handle low-quality and missing modalities in a unified framework. 

\textbf{(1).} For samples with missing modalities, we recover the missing modalities from the available ones by supervising the recovery process based on samples and distributions, thereby encoding them the same as complete samples. The sample-based recovery forces the model to convert between image and text features of the same samples. It effectively builds local connections between modalities, yet is prone to be misled by the mismatches of image-text pairs. Therefore, we introduce an additional distribution-based recovery, facilitating conversion between image and text distributions. Concretely, it encourages the model to predict the mean and variance of one distribution from another. This provides global mapping relationships between modalities and eliminates the negative impacts of the mismatches. 

\textbf{(2).} For samples with diverse-quality modalities, we determine the contribution of each modality to the fusion based on its correlation with the distribution. Leveraging the global mapping relationships learned by the modality recovery process, we use the recovered modalities that conform to the distributions to compensate for potential low-quality modalities and expand each sample into three. Then, we quantitatively estimate the quality of each modality with Gaussian distribution probability and assign weights for three samples by multiplying the probabilities of its two source modalities. Finally, we compute the overall fused feature as the weighted sum of the three fused features. Through this process, we can dynamically fuse modalities according to their qualities, reducing the influences of low-quality modalities on the fusion.


To systematically assess the robustness of models, we adopt two disruption strategies that randomly corrupt and discard modalities from samples to mimic real-world scenarios of various degrees of low-quality and missing modalities. By conducting extensive experiments on MVSA-S, MVSA-M \cite{mmm2016mvsa}, and TumEmo \cite{tmm2021mvan}, we prove the effectiveness of DRF in robust MSA. The main contribution of this paper is summarized as follows:

\begin{itemize}[leftmargin=*]
\item We focus on robust MSA of image-text pairs for the low-quality and missing modalities, which are prevalent concerns in real-world scenarios. As far as we know, this is the first attempt to explore the robustness of models in this subfield.

\item We propose a novel method, DRF, to handle the low-quality and missing modalities in a unified framework. It leverages two feature distributions to provide global mapping relationships between modalities for feature recovery as well as qualitative estimations of modality quality for feature fusion.  

\item Experimental results under two disruption strategies on three MSA benchmark datasets demonstrate the significant improvements of DRF compared to the state-of-the-art MSA methods, validating its superiority in robust MSA of image-text pairs.

\end{itemize}

\section{Related Works}
\subsection{Multimodal Sentiment Analysis}
Early works on sentiment analysis focus solely on a single modality, such as text \cite{ftir2007text1, cl2011text2},  image \cite{icip2008image1, aaai2015image2} and speech \cite{tsap2012speech1, isca2017speech2}. With the rapid increase of posts in social media, MSA for image-text pairs has garnered increasing attention in recent years. In the beginning, researchers leverage the semantics of images and texts with simple concatenation \cite{isi2017hsan} or attention \cite{cikm2017multisentibank}. Later on, more elaborate attention-based structures are designed to enable more comprehensive modality fusion. COMN \cite{sigir2018comem} iteratively models the interaction between image and text features at multiple levels. MVAN \cite{tmm2021mvan} fully exploits the correlations of different views of images and texts. CLMLF \cite{naacl2022clmlf} leverages Transformer-Encoder \cite{nips2017transformer} for token-level alignments. More recently, the focus of researchers has shifted toward addressing task-specific challenges. MGNNS \cite{acl2021mgnns} utilizes graph neural networks to capture the global characteristics of the dataset. MVCN \cite{acl2023mvcn} tackles the modality heterogeneity with sparse attention, feature restraint, and loss calibration. UP-MPF \cite{mm2022upmpf} and MultiPoint \cite{mm2023multipoint} devote to few-shot MSA to avoid annotation costs. There is also a series of studies \cite{ijcai2019tombert, mm2021captrbert, emnlp2022fite, acl2022vlpmabsa} on fine-grained MSA, aiming to detect the sentiment of a specific aspect within the image-text pair, which though is not the primary focus of this paper.

These methods effectively model the sentiments by relying on complementary information from both images and texts, yet can not properly handle the issues of low-quality and missing modalities. Since these issues might frequently occur in real-life applications \cite{nips2019cpmnet}, we propose DRF, a practical method capable of predicting sentiment for image-text pairs robustly.

\begin{figure*}
  \centering
  \includegraphics[width=1\textwidth]{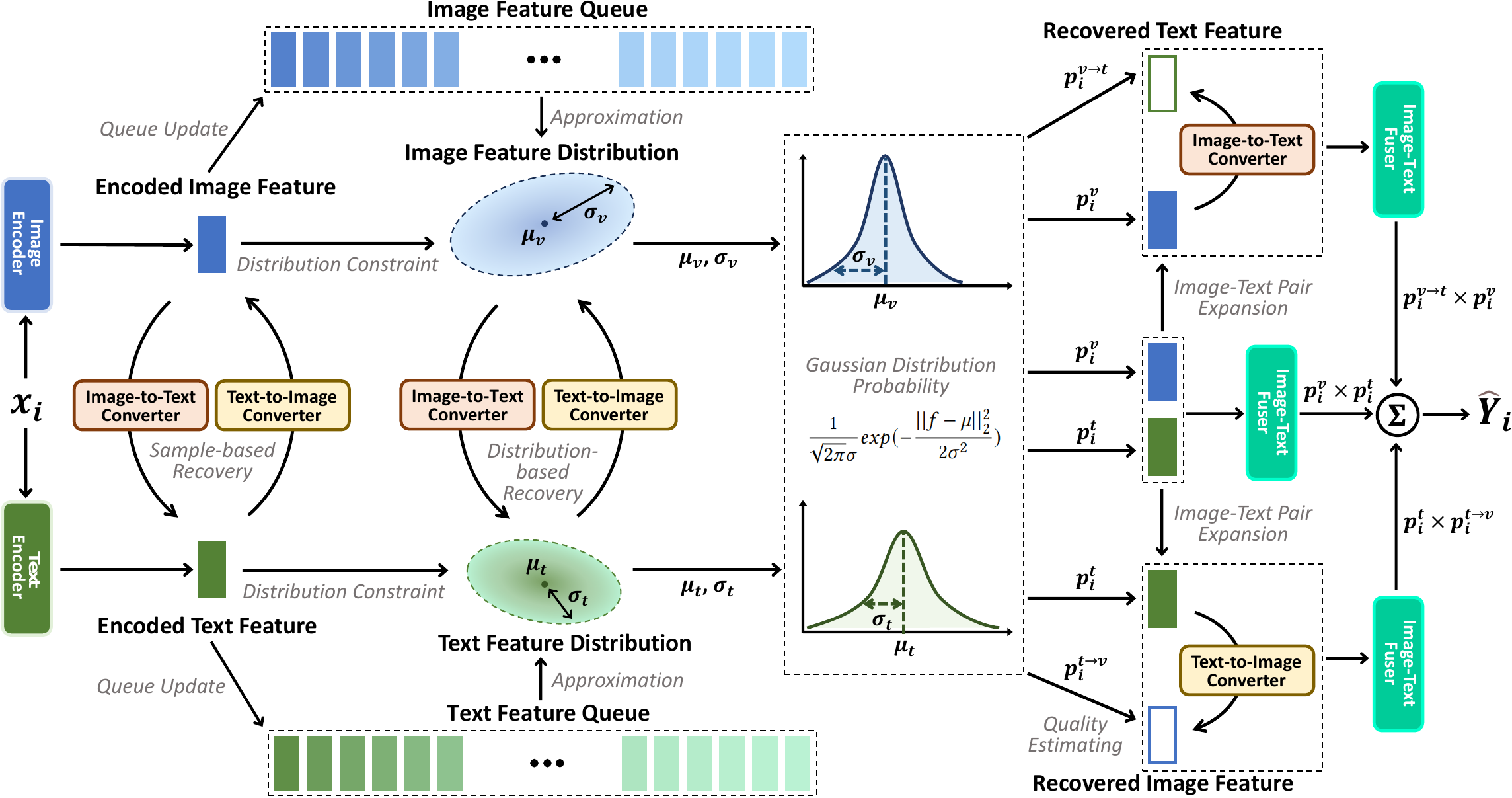}
  \vspace{-15pt}
  \caption{Illustration of DRF. The core of our method is the modeling of image and text feature distributions, which we approximate using the respective feature queues. After separate encoding of each modality, we first supervise two converters to learn inter-modal mapping relationships by sample-based and distribution-based recovery. Subsequently, we leverage the recovered features to expand each sample into three. Utilizing the Gaussian distribution probability, we estimate the modality qualities to decide their contributions to the fusion. Finally, we obtain the overall fused feature as the weighted sum of the features of three expanded samples and enqueue features to the queue according to their qualities. }
  \label{framework}
  \vspace{-5pt}
\end{figure*}

\subsection{Robust Multimodal Learning}
The issues of low-quality and missing modalities are prevalent in all types of multimodal data, and various studies have been conducted on them. For low-quality modalities, a feasible strategy is to reduce their influences on the fusion as adopted in trusted multi-view classification \cite{iclr2021tmvc, pami2022tmvc}. Researchers estimate the uncertainty of each view based on Dempster-Shafer Evidence Theory \cite{1967dst, shafer1976dst} and give less consideration to the high uncertainty views, which correspond to the low-quality modalities in our case, during the fusion. The uncertainty is also estimated according to other methods or theories in related studies, including Bayesian neural networks \cite{nips1990bnn, icml2016dropout}, ensemble-based methods \cite{nips2017ensemble, iclr2021ensemble2}, Normal Inverse-Gamma distribution \cite{nips2021monig} and energy score \cite{nips2020energy1, icml2023qmf}. For missing modalities, data imputation methods \cite{pami2023gcnet} in incomplete multimodal learning recover them from the available ones. To achieve this, some researchers directly pad missing modalities with fixed values \cite{pami2022fixpad1, kdd2020fixpad2}, some others optimize through low-rank projection \cite{siam2010low-rank1, jmlr2010low-rank2}, the rest leverage the generative capability of specific neural networks architectures, such as autoencoder \cite{icml2008autoencoder} and Transformer \cite{nips2017transformer}.

To unifiedly handle both issues in MSA of image-text pairs, we leverage the image and text feature distributions. On the one hand, the distributions can provide quantitative estimations of modality qualities through the probability density function. On the other hand, they can also guide the learning of global mapping relationships between modalities, eliminating the negative impacts of image-text pair mismatches.

\section{Method}

\subsection{Task Formulation}
\label{task formulation}
We focus on the sentiment classification of image-text pairs with possible low-quality and missing modalities. We first give a definition of the regular MSA. Given a set of samples $\{(x_i,y_i)|i\in \{1,2,\cdots, N\}\}$, where $x_i$ denotes the image-text pair $(v_i, t_i)$, $y_i$ is its sentiment label from a total of $S$ categories, and $N$ is the total number of samples, the model needs to build a mapping between image-text pairs $\boldsymbol{x}$ and sentiment labels $\boldsymbol{y}$.

To simulate the occurrences of low-quality and missing modalities in real-world applications, we randomly corrupt and discard modalities from samples. We denote the discarding operation of image-text pair $(v_i, t_i)$ as $\lambda^v_i, \lambda^t_i \in \{0,1\}$. Take image $v_i$ as an example: $\lambda_i^v=0$ represents that it is discarded, in other words, missing, and $\lambda_i^v=1$ represents the other way. For the corruption operation aimed at simulating low-quality modalities, we consider it invisible to the model because it is also difficult to accurately pre-determine modality quality in practice. Thus, the overall definition of $x_i$ in robust MSA is $(v_i, t_i, \lambda_i^v, \lambda_i^t)$.

\subsection{Feature Distribution Modeling}
\label{feature encoding}
The pipeline of DRF is shown in \cref{framework}. For convenience, we pretend both the image and text are not discarded while presenting our method and reflect the influences of $\lambda_i^v, \lambda_i^t$ by the computations. After receiving the image-text pair $x_i = (v_i, t_i, \lambda_i^v, \lambda_i^t)$ of an input sample $(x_i,y_i)$, we first encode $v_i$ into image feature $f_i^v \in \mathbb{R}^{d_v}$, and $t_i$ into text feature $f_i^t \in \mathbb{R}^{d_t}$. $d_v,d_t$ are the feature dimensions of the image and text. 

In our framework, the core of unified modeling of low-quality and missing modalities is the feature distribution of each modality. To acquire these distributions, limited features from a single mini-batch are insufficient. Inspired by self-supervised learning \cite{cvpr2018id, cvpr2020moco}, we maintain a feature queue for each modality to record features across multiple mini-batches. The feature queue of image is denoted by $Q_v = \{f_j^v|j\in q_v\}$ and it of text is denoted by $Q_t = \{f_j^t|j\in q_t\}$, with the queue size set to $L$ for both of them. By adopting a sufficiently large queue size, we can approximate the feature distributions of all samples by those from feature queues. Specifically, we approximate the mean $\mu_v$ and standard deviation $\sigma_v$ of the image feature distribution by:
\begin{equation}
\label{image distribution1}
\mu_v = \frac{1}{L}\sum_{j\in q_v}f^v_j,
\end{equation}
\begin{equation}
\label{image distribution2}
\sigma_v = \sqrt{\frac{1}{L}\sum_{j\in q_v}{||f^v_j-\mu_v||_2^2}}.
\end{equation}
The mean $\mu_t$ and standard deviation $\sigma_t$ of the text feature distribution are approximated similarly. 

To encourage the compactness of each distribution and the separation between distributions, we devise a distribution constraint that brings image and text features closer to the means of their respective feature distributions and away from the means of the other:
\begin{equation}
\label{compact}
\begin{aligned}
\mathcal{L}_{dis} &= \lambda_i^v \cdot exp(||f^v_i-\mu_v||_2-||f^v_i-\mu_t||_2) 
\\&+ \lambda_i^t \cdot exp(||f^t_i-\mu_t||_2-||f^t_i-\mu_v||_2).
\end{aligned}
\end{equation}

\subsection{Modality Recovery}
\label{inter-modal recovery}
To handle missing modalities, we build mapping relationships between image and text through two modality converters, which are essentially two-layer MLPs. For the image-to-text converter, denoted by $\boldsymbol{C}_{v\to t}(\cdot)$, an intuitive idea is encouraging it to recover the text feature $f_i^t$ from the image feature $f_i^v$. We call this task sample-based recovery and its loss is given by:
\begin{equation}
\label{sample-based recovery}
\mathcal{L}_{v\to t}^{s} = \lambda_i^v\lambda_i^t \cdot ||\boldsymbol{C}_{v\rightarrow t}(f_i^v) - f_i^t||_2.
\end{equation}
Its effectiveness is built upon the alignment between information of image $v_i$ and text $t_i$. However, due to the mismatches between images and texts from social media posts \cite{acl2021mgnns}, such alignment can not be guaranteed for all samples, leading to occasionally negative impacts on the converter. To alleviate these, we devise a distribution-based recovery task that provides mapping guidance from a global perspective. Specifically, we supervise the converter to recover the mean $\mu_t$ and standard deviation $\sigma_t$ of $Q_t$ from $Q_v$. The mean $\mu_{v\to t}$ and standard deviation $\sigma_{v\to t}$ of the converted distribution are computed as:
\begin{equation}
\label{recovered text distribution1}
\mu_{v\to t} = \frac{1}{L}\sum_{j\in q_v}\boldsymbol{C}_{v\to t}(f^v_j),
\end{equation}
\begin{equation}
\label{recovered text distribution2}
\sigma_{v\to t} = \sqrt{\frac{1}{L}\sum_{j\in q_v}{||\boldsymbol{C}_{v\to t}(f^v_j)-\mu_{v\to t}||_2^2}}.
\end{equation}
Then, the loss of distribution-based recovery is given by:
\begin{equation}
\label{distribution-based recovery}
\mathcal{L}_{v\to t}^{d} = ||\mu_{v\to t}-\mu_t||_2 + |\sigma_{v\to t}-\sigma_t|.
\end{equation}
The sample-based and distribution-based recovery tasks are also applied to the text-to-image converter $C_{t\to v}(\cdot)$ with symmetric computations. Thereby, the combined loss of both converters is:
\begin{equation}
\label{overall recovery}
\mathcal{L}_{rec} = \mathcal{L}_{v\to t}^{s} + \mathcal{L}_{v\to t}^{d} + \mathcal{L}_{t\to v}^{s} + \mathcal{L}_{t\to v}^{d}.
\end{equation}

\subsection{Modality Quality Estimation}
\label{fusion and classification}
To handle samples with potentially low-quality modalities, we perform multimodal fusion based on the quality of each modality estimated by the feature distributions. Firstly, we expand the image-text pair into three, by treating its image $v_i$ and text $t_i$ as independent samples with missing modalities. Through the modality recovery process, we obtain the recovered image feature $\boldsymbol{C}_{t\to v}(f^t_i)$, denoted by $f^{t\to v}_i$ and the recovered text feature $\boldsymbol{C}_{v\to t}(f^v_i)$, denoted by $f^{v\to t}_i$. Thus, the image and text features of the original sample are $(f^v_i, f^t_i)$, those of the image are $(f^v_i, f^{v\to t}_i)$, and those of the text are $(f^{t\to v}_i, f^t_i)$.

\begin{figure}[t]
      \centering
      \includegraphics[width=1\linewidth]{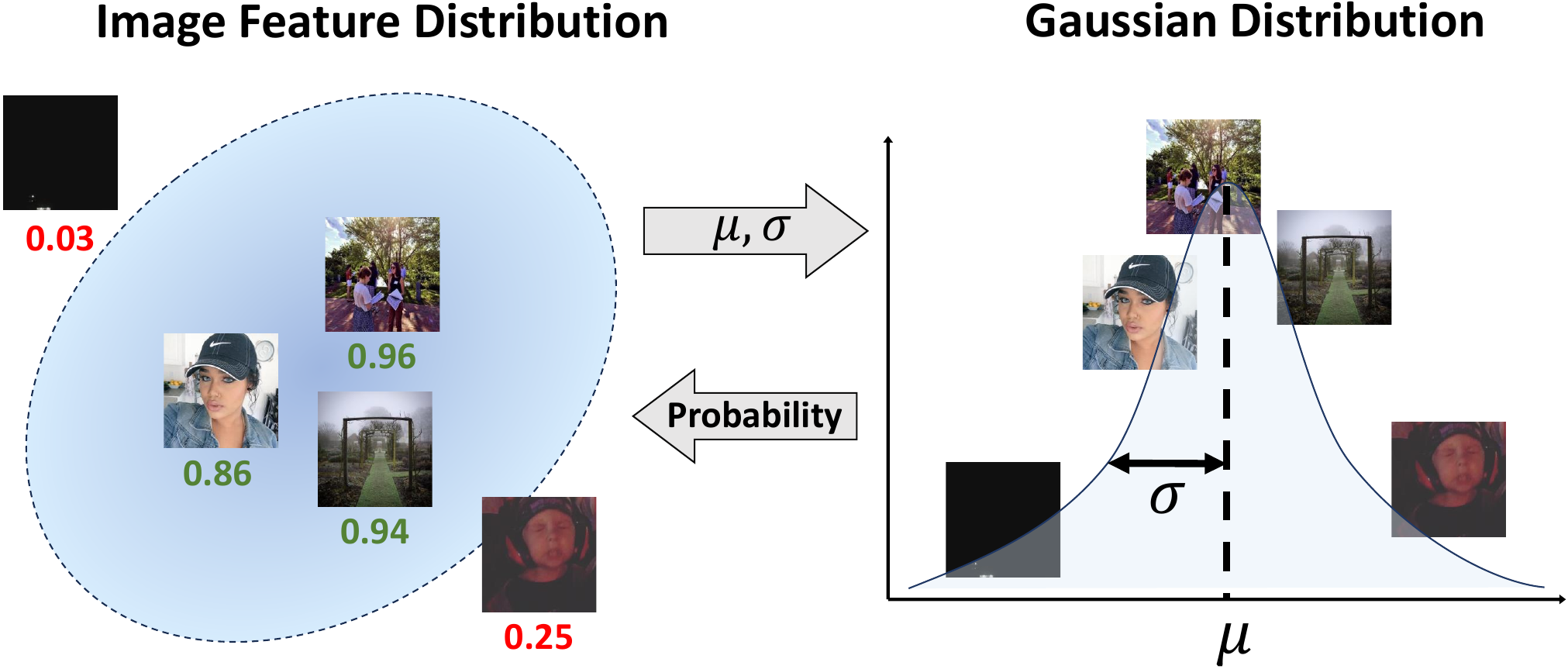}
      \vspace{-6pt}
      \caption{Examples of estimating image quality based on the feature distribution.}
      \label{quality illustration}
\end{figure}

Subsequently, we estimate the quality of each modality according to its correlation with the respective feature distribution. We consider those unimodal features that conform to the feature distribution to come from high-quality modalities, while the others to come from low-quality modalities. We adopt the Gaussian distribution to provide quantitative estimations. Its probability density function given feature $f$, mean $\mu$ and standard deviation $\sigma$ is:
\begin{equation}
\label{gaussian distribution}
p(f, \mu, \sigma) = \frac{1}{\sqrt{2\pi}\sigma}exp(-\frac{||f-\mu||^2_2}{2\sigma^2}).
\end{equation}
We compute the contributions of $f^v_i$ and $f^{t\to v}_i$ to the fusion as the probabilities of them belonging to the image feature distribution:
\begin{equation}
\label{compute visual prob}
p^v_i = p(f^v_i, \mu_v, \sigma_v),\quad p^{t\to v}_i = p(f^{t\to v}_i, \mu_v, \sigma_v),
\end{equation}
and the contributions of $f^t_i$ and $f^{v\to t}_i$ to the fusion as the probabilities of them belonging to the text feature distribution:
\begin{equation}
\label{compute textual prob}
p^t_i = p(f^t_i, \mu_t, \sigma_t),\quad p^{v\to t}_i = p(f^{v\to t}_i, \mu_t, \sigma_t).
\end{equation}
A few examples are demonstrated in \cref{quality illustration} for illustration. Then, we fuse the image and text features of each sample by feeding them into a shared three-layer MLP $\boldsymbol{F}_{v+t}(\cdot)$ after concatenation and obtain the overall fused feature $M_i$ by the weighted sum.
\begin{equation}
\begin{aligned}
\label{overall fusion}
M_i &= \lambda_i^v\lambda_i^t \cdot (p^v_i p^t_i) \cdot \boldsymbol{F}_{v+t}([f^v_i, f^t_i])
\\&+ \lambda_i^v \cdot (p^v_i p^{v\to t}_i) \cdot \boldsymbol{F}_{v+t}([f^v_i, f^{v\to t}_i])
\\&+ \lambda_i^t \cdot (p^{t\to v}_i p^t_i) \cdot \boldsymbol{F}_{v+t}([f^{t\to v}_i, f^t_i]).
\end{aligned}
\end{equation}
Through this process, we explicitly reduce the contributions of low-quality modalities to the fusion, enabling reliable fusion for potential low-quality modalities.


During training, the parameters of encoders are gradually changing, resulting in smooth shifting of the feature distributions. To track it, we need to progressively update the feature queues with the features from the latest encoders. Meanwhile, we hope to retain the capability of the feature distributions to distinguish modalities of different qualities. To satisfy both requirements, we update the queues with the encoded features of the current sample that exhibit correlations with their respective feature distributions. Specifically, take image $v_i$ as an example, we enqueue $p^v_i$ to $Q_v$ if its probability of belonging to the image feature distribution is larger than the mean of the probabilities of features in $Q_v$:
\begin{equation}
\label{update}
p^v_i > \frac{1}{L}\sum_{j\in q_v}{p(f^v_j, \mu_v, \sigma_v)}.
\end{equation}
The update strategy for the text feature queue $Q_t$ is similar.

\subsection{Classification and Optimization}

For sentiment prediction, we feed the overall fused feature $M_i$ into a fully connected layer followed by a softmax layer:
\begin{equation}
\label{prediction1}
\hat{Y_i} = softmax(WM_i+b),
\end{equation}
where $W, b$ are trainable parameters of the fully connected layer, $\hat{Y_i}$ is the predicted probabilities of $S$ sentiment categories. We denote the predicted probability for $k$-th category as $\hat{y}_i^k$, and constrain the classification by a cross-entropy loss:
\begin{equation}
\label{prediction2}
\mathcal{L}_{cls} = -\sum^{S}_{k=1} y_i log(\hat{y}_i^k).
\end{equation}
To this end, the joint optimization objective for all parameters is:
\begin{equation}
\label{overall}
\mathcal{L} = \mathcal{L}_{dis} + \mathcal{L}_{rec} + \mathcal{L}_{cls}.
\end{equation}

\begin{table*}[t]
  \caption{Model performances under modality-fixed disruption. We report the ACC/F1 scores of models under C, D, and C+D settings on MVSA-S, MVSA-M, and TumEmo. The highest result is highlighted in bold.}
  \vspace{-6pt}
  \centering
  \renewcommand\arraystretch{1.1}
  \resizebox{1\linewidth}{!}{
  \begin{tabular}{c|l|ccc|ccc|ccc}
    \cline{1-11}
    \textbf{Disrupted} & \multirow{2}*{\textbf{Method}} & \multicolumn{3}{c|}{\textbf{MVSA-S}} & \multicolumn{3}{c|}{\textbf{MVSA-M}} & \multicolumn{3}{c}{\textbf{TumEmo}} \\
    \cline{3-11}
    \textbf{Modality}& & \textbf{C} & \textbf{D} & \textbf{C+D} & \textbf{C} & \textbf{D} & \textbf{C+D} & \textbf{C} & \textbf{D} & \textbf{C+D} \\
    \cline{1-11}
    \multirow{6}*{\textbf{Image}} & HSAN \cite{isi2017hsan} & 70.5/69.7 & 69.8/69.6 & 70.0/69.5 & 67.5/65.6 & 66.2/64.1 & 66.6/64.3 & 63.5/63.3 & 62.5/62.4 & 62.9/62.8 \\
    & MVAN \cite{tmm2021mvan} & 67.7/67.4 & 66.5/66.0 & 66.3/66.2 & 66.9/64.8 & 66.0/63.7 &66.4/64.2 & 60.7/60.6 & 60.1/60.0 & 60.4/60.4 \\
    & MGNNS \cite{acl2021mgnns} & 71.9/71.8 & 71.6/70.9 & 71.6/71.3 & 69.4/66.3 & 68.6/65.7 & 69.1/66.2 & 65.2/65.1 & 63.8/63.6 & 64.1/64.0 \\
    & CLMLF \cite{naacl2022clmlf} & 69.4/69.0 & 67.7/67.8 & 68.4/68.1 & 67.0/65.3 & 66.4/64.3 & 66.7/65.0 & 62.4/62.3 & 61.8/61.5 & 62.2/62.1 \\
    & MVCN \cite{acl2023mvcn} & 70.3/69.9 & 69.3/69.2 & 69.9/69.4 & 68.1/66.0 & 67.3/64.9 & 67.6/65.3 & 63.7/63.6 & 62.9/62.9 & 63.3/63.3 \\
    & \textbf{DRF} (\textbf{Ours}) & \textbf{74.5/74.4} & \textbf{73.4/73.1} & \textbf{73.8/73.6} &\textbf{71.0/68.2} & \textbf{70.0/67.5} & \textbf{70.3/67.9} & \textbf{68.4/68.2} & \textbf{67.2/67.2} & \textbf{67.9/67.7}\\
    \cline{1-11}
    \multirow{6}*{\textbf{Text}} & HSAN \cite{isi2017hsan} & 64.9/64.3  & 64.1/63.3 & 64.6/64.2 & 64.4/61.6 & 62.9/60.7 & 63.6/61.4 & 48.8/48.5 & 47.5/47.4 & 48.2/48.0\\ 
    & MVAN \cite{tmm2021mvan} & 63.0/62.3 & 62.4/62.2 & 62.8/62.5 & 64.1/60.9 & 62.9/60.0 & 63.5/61.7 & 45.3/45.2 & 44.4/44.0 & 44.8/44.7 \\
    & MGNNS \cite{acl2021mgnns} & 66.1/65.6 & 64.7/64.5 & 65.5/65.2 & 64.8/62.5 & 63.5/61.8 & 64.1/62.3 & 52.6/52.7 & 50.4/50.4 & 51.5/51.3 \\
    & CLMLF \cite{naacl2022clmlf} & 64.3/63.6 & 63.1/62.8 & 63.7/63.4 & 63.8/61.2 & 62.5/60.4 & 63.3/60.7 & 48.1/48.0 & 46.9/46.7 & 47.0/46.9 \\
    & MVCN \cite{acl2023mvcn} & 65.3/65.0 & 64.6/64.5 & 65.0/64.7 & 64.4/62.1 & 63.3/61.4 & 63.8/61.9 & 50.5/50.3 & 49.2/49.2 & 49.8/49.7 \\
    & \textbf{DRF} (\textbf{Ours}) & \textbf{69.4/69.4} & \textbf{68.1/68.0} & \textbf{68.5/68.3} & \textbf{67.9/66.5} & \textbf{67.2/64.8} & \textbf{67.3/66.2} & \textbf{61.6/61.4} & \textbf{59.2/59.1} & \textbf{60.9/61.0} \\
    \cline{1-11}
  \end{tabular}}
  \label{fixed comparison}
\end{table*}

\begin{table}[t]
  \caption{Statistics of datasets.}
  \vspace{-5pt}
  \centering
  \resizebox{0.8\linewidth}{!}{
  \begin{tabular}{cccccc}
    \toprule
    \textbf{Dataset} & \textbf{Total}& \textbf{Train}& \textbf{Val} & \textbf{Test}\\
    \midrule
    MVSA-S \cite{mmm2016mvsa} & 4511 & 3608  & 451   & 452   \\
    MVSA-M \cite{mmm2016mvsa} & 17024 & 13618 & 1703  & 1703  \\
    TumEmo \cite{tmm2021mvan} & 195265 & 156217 & 19524 & 19524  \\
    \bottomrule
  \end{tabular}}
  \vspace{-6pt}
  \label{statistics}
\end{table}

\section{Experiment}
\subsection{Dataset Preparations}
We carry out experiments on three publicly available MSA datasets. The statistics of them are presented in \cref{statistics}. \textbf{MVSA-S} and \textbf{MVSA-M} \cite{mmm2016mvsa} are two Twitter datasets annotated by sentiment polarities: \{\textit{positive, neutral, negative}\}. We pre-process their samples following Xu and Mao \cite{cikm2017multisentibank}. \textbf{TumEmo} \cite{tmm2021mvan} is a Tumblr dataset annotated according to the emotions of tags. It has 7 emotion categories: \{\textit{angry, bored, calm, fear, happy, love, sad}\}. We follow the pre-processing of Yang \textit{et al.} \cite{tmm2021mvan} for a fair comparison. We report the accuracy score (\textbf{ACC}) and F1 score (\textbf{F1}) for all three datasets.

\begin{figure}[b]
      \centering
      \vspace{-6pt}
      \includegraphics[width=1\linewidth]{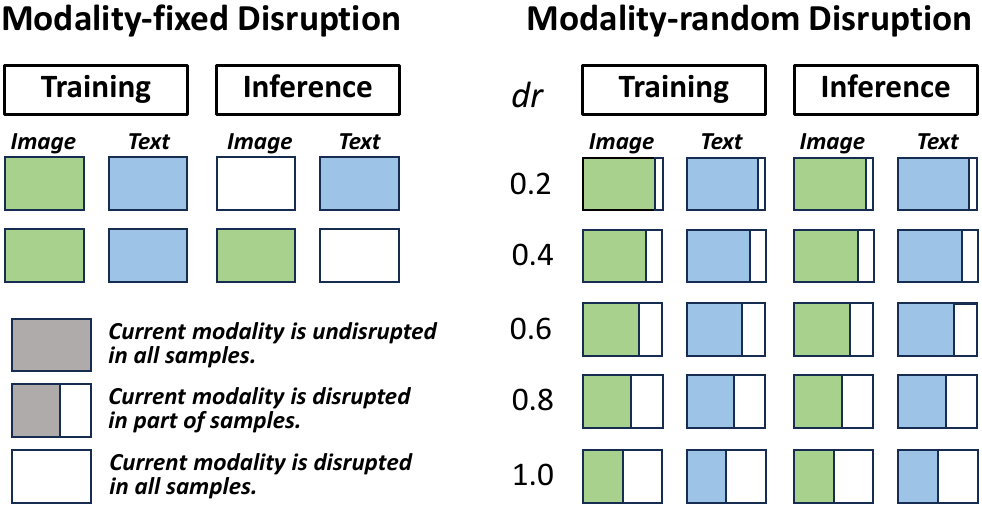}
      \caption{Illustration of modality-fixed disruption and modality-random disruption strategies.}
      \vspace{-6pt}
      \label{disruption illustration}
\end{figure}

To evaluate the robustness of models to low-quality and missing modalities, we simulate these cases by performing two kinds of disruptions on samples. To simulate low-quality modalities, we corrupt images by randomly masking 40-80\% of pixels, and texts by replacing 40-80\% of words with [MASK] tokens. To simulate missing modalities, we discard modalities from samples. By referring to related fields \cite{aaai2019cycle-translate, acl2021mmin, nips2023incomplete}, we incorporate two disruption strategies for a systematical evaluation: modality-fixed disruption and modality-random disruption. In \textbf{modality-fixed disruption}, we do not interfere with the training process and disrupt a fixed modality for all samples during inference. In \textbf{modality-random disruption}, we disrupt a random modality for a pre-defined ratio of samples in both training and inference. At least one modality in each sample is guaranteed to be undisrupted, and reliable for the sentiment prediction. We use the disruption ratio ($\boldsymbol{dr}$) to represent the ratio of samples disrupted and conduct experiments for $\boldsymbol{dr} \in \{0.2,0.4,0.6,0.8,1.0\}$. We illustrate the two strategies in \cref{disruption illustration}. For each strategy, we investigate three settings: only corrupts modalities (\textbf{C}), corresponding to only introducing low-quality modalities; only discards modalities (\textbf{D}), corresponding to only introducing missing modalities; and corrupts and discards modalities half-to-half (\textbf{C+D}), corresponding to introducing both low-quality and missing modalities. 

\subsection{Implementation Details}
For the image encoder, we adopt Vision Transformer \cite{iclr2021vit} with a patch size of 16, and resize images to 224 $\times$ 224. The obtained image features are $d_v=768$ dimensions. For text, we adopt Bert \cite{naacl2019bert} to obtain text features with the same $d_t=768$ dimensions. These settings are consistent with the recent SOTA method MVCN \cite{acl2023mvcn} for a fair comparison. We set the mini-batch size to 16 and queue size $L$ to 512. We train the model for 30 epochs with AdamW optimizer. The initial learning rate is set to 2e-5 for image and text encoders and 2e-4 for the rest of the parameters. The learning rates are decayed to 1e-6 in the cosine schedule.

\begin{figure*}[t]
      \centering
      \includegraphics[width=0.95\linewidth]{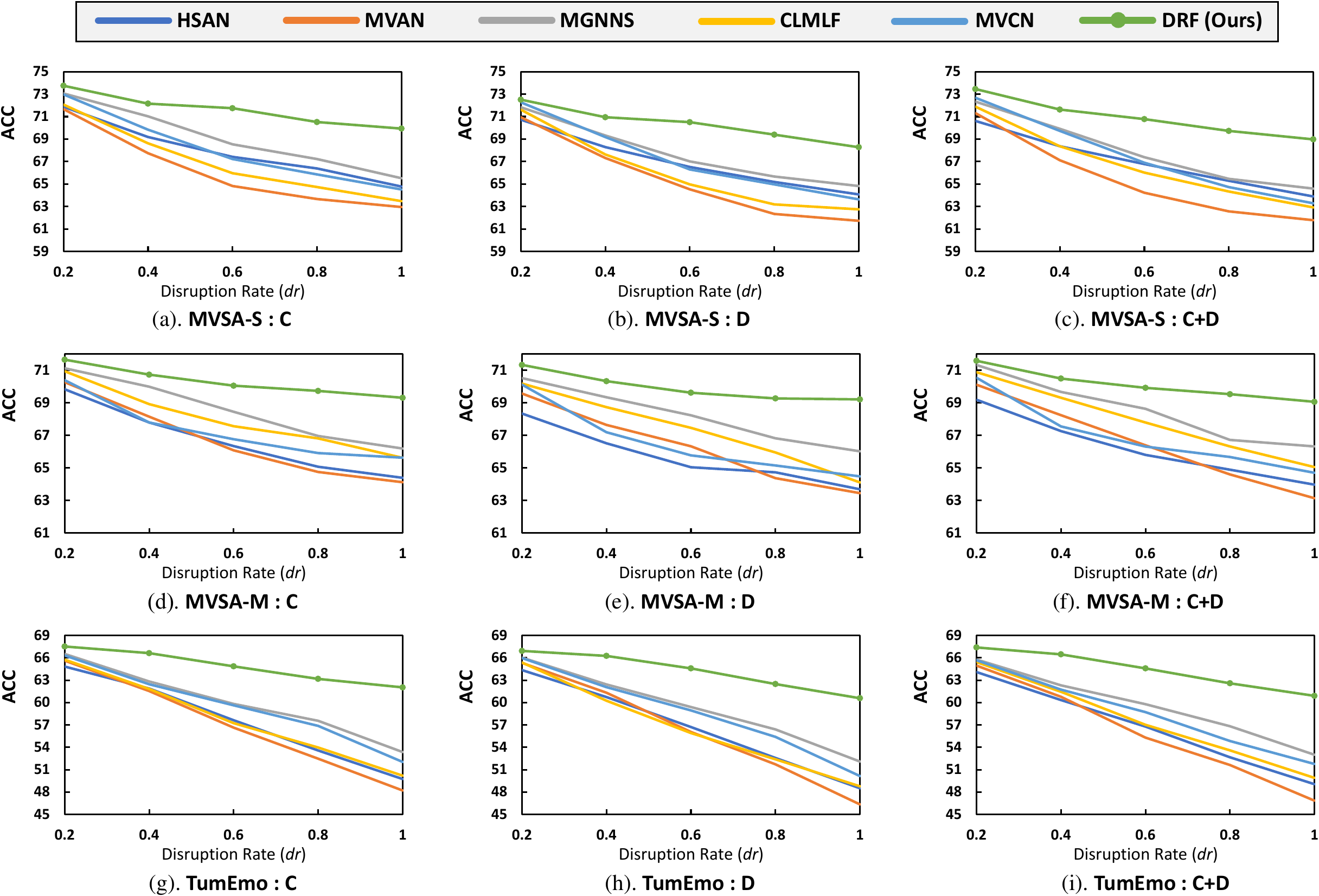}
      \vspace{-6pt}
      \caption{Model performances under modality-random disruption. We report ACC scores of models under C, D, and C+D settings on MVSA-S, MVSA-M, and TumEmo.}
      \label{random comparison}
      \vspace{-6pt}
\end{figure*}

\subsection{Compared Methods}
We compare DRF with a series of SOTA MSA methods to comprehensively validate its effectiveness in robust sentiment classification of image-text pairs. We present brief introductions for the compared methods below. For methods incapable of receiving input with missing modalities, we pad images with blank pixels and texts with [MASK] tokens.

\textbf{HSAN} \cite{isi2017hsan} employs image captions to extract image features and concatenates them with text features for sentiment prediction. We reproduce it by replacing its text encoder with a more advanced BERT \cite{naacl2019bert}. 

\textbf{MVAN} \cite{tmm2021mvan} separately encodes the object and scene features in images, and interactively models their dependencies with the text features through a memory network.

\textbf{MGNNS} \cite{acl2021mgnns} first introduces graph neural network into MSA, which captures the global co-occurrence characteristics in texts and images, enabling global-aware modality fusion.

\textbf{CLMLF} \cite{naacl2022clmlf} fuses modalities based on Transformer-Encoder \cite{nips2017transformer} to facilitate token-level alignments between modalities. It also proposes two contrastive learning tasks aiding in learning common sentiment features.

\textbf{MVCN} \cite{acl2023mvcn} tackles the modality heterogeneity from three views: (1). it proposes a sparse attention mechanism to filter out redundant visual features; (2). it restrains representations to calibrate the feature shift; (3) it alleviates the uncertainty in annotations through an adaptive loss calibration.

\subsection{Comparision with the State-Of-The-Art}
\subsubsection{\textbf{Modality-fixed Disruption}}
The comparison under the strategy of modality-fixed disruption is displayed in \cref{fixed comparison}. DRF consistently achieves the highest results across all cases. It indicates that compared with current methods, DRF is more robust to both low-quality and missing modalities through explicit modeling of modality qualities and building inter-modal mapping relationships. The advantages of DRF under the disruption of texts are more significant. We conjecture that other methods depend more on texts than images due to the higher information density of texts \cite{acl2019mult}. Subsequently, the corruption and discarding of texts results in severe degeneration of their performances. In contrast, DRF alleviates those influences by flexibly adjusting the contribution of texts and recovering the absent text feature.

\begin{table}[t]
\centering
\renewcommand\arraystretch{1.1}
\caption{Model performances without disruption. We report ACC/F1 scores of models on MVSA-S, MVSA-M, and TumEmo. The highest result is highlighted in bold, and the second-highest result is underlined.}
\vspace{-6pt}
\resizebox{0.8\linewidth}{!}{
\begin{tabular}{l|c|c|c}
\toprule
\textbf{Method} & \textbf{MVSA-S} & \textbf{MVSA-M} & \textbf{TumEmo} \\
\midrule
HSAN \cite{isi2017hsan} & 69.9/66.9 &68.0/67.8 &63.1/54.0\\
MVAN \cite{tmm2021mvan} & 73.0/73.0 & \underline{72.4}/\textbf{72.3}  & 66.5/63.4 \\
MGNNS \cite{acl2021mgnns} & 73.8/72.7 &\textbf{72.5}/69.3 &66.7/66.7\\
CLMLF \cite{naacl2022clmlf} & 75.3/73.5 &71.1/68.6 & 68.1/68.0\\
MVCN \cite{acl2023mvcn} & \underline{76.1}/\underline{74.6} &72.1/70.0 & \underline{68.4}/\underline{68.4}\\
\textbf{DRF} (\textbf{Ours}) & \textbf{76.5/75.9} &72.2/\underline{70.4} &\textbf{69.6/69.6} \\
\bottomrule
\end{tabular}}
\vspace{-6pt}
\label{without disruption}
\end{table}

\begin{table*}[t]
  \centering
  \renewcommand\arraystretch{1.1}
  \caption{Ablation study of components under modality-fixed disruption on MVSA-S and TumEmo. Sample-based recovery and distribution-based recovery are the two kinds of supervision on the modality converters introduced in \cref{inter-modal recovery}. Gaussian distribution probability is adopted to estimate the quality of modalities. Image-text expansion is the process of expanding each sample into three. They are from \cref{fusion and classification}. Distribution constraint encourages the compactness in feature distributions and separation between feature distributions, computed by \cref{compact}. Experiments for separate components are conducted independently.}
  \resizebox{1\linewidth}{!}{
  \begin{tabular}{c|l|ccc|ccc}
    \cline{1-8}
    \textbf{Disrupted} & \multirow{2}*{\textbf{Method}} & \multicolumn{3}{c|}{\textbf{MVSA-S}} & \multicolumn{3}{c}{\textbf{TumEmo}} \\
    \cline{3-8}
    \textbf{Modality}& & \textbf{C} & \textbf{D} & \textbf{C+D} & \textbf{C} & \textbf{D} & \textbf{C+D} \\
    
    \cline{1-8}
    \multirow{6}*{\textbf{Image}} & \textbf{DRF} & \textbf{74.5/74.4} & \textbf{73.4/73.1} & \textbf{73.8/73.6} & \textbf{68.4/68.2} & \textbf{67.2/67.2} & \textbf{67.9/67.7}\\
    & w/o Sample-based Recovery & 73.7/73.4 & 72.1/72.0 & 72.6/72.1 & 68.1/67.9 & 66.0/65.9 & 67.0/67.0 \\
    & w/o Distribution-based Recovery & 73.2/72.5 & 71.5/71.2 & 72.2/71.6 & 67.7/67.6 & 65.5/65.6 & 66.7/66.6 \\
    & w/o Gaussian Distribution Probability & 71.9/71.7 & 72.7/72.3 & 72.3/72.1 & 65.0/64.7 & 66.6/66.7 & 65.8/65.8  \\
    & w/o Image-text Pair Expansion & 72.4/72.2 & 68.3/67.1 & 71.0/70.6 & 66.6/66.5 & 62.8/62.7 & 64.6/64.4 \\
    & w/o Distribution Constraint & 74.0/73.8 & 72.5/72.0 & 73.5/73.2 & 67.9/67.9 & 66.3/66.2 & 67.1/67.1 \\
    \cline{1-8}
    \multirow{6}*{\textbf{Text}} & \textbf{DRF} & \textbf{69.4/69.4} & \textbf{68.1/68.0} & \textbf{68.5/68.3} & \textbf{61.6/61.4} & \textbf{59.2/59.1} & \textbf{60.9/61.0} \\
    & w/o Sample-based Recovery & 68.5/68.3 & 66.7/66.4 & 67.5/66.8 & 60.2/60.0 & 57.8/57.7 & 58.9/58.8  \\
    & w/o Distribution-based Recovery & 68.5/68.4 & 65.8/65.2 & 67.0/66.9 & 60.4/60.4 & 57.5/57.6 & 59.0/58.8 \\
    & w/o Gaussian Distribution Probability & 67.1/66.7 & 67.5/67.5 & 67.3/67.0 & 58.8/58.7 & 58.4/58.4 & 58.6/58.6 \\
    & w/o Image-text Pair Expansion & 67.7/67.2 & 65.0/64.8 & 66.2/65.9 & 59.3/59.2 & 53.1/53.0 & 56.2/56.3 \\
    & w/o Distribution Constraint & 68.7/68.5 & 67.2/67.0 & 67.9/67.6 & 61.3/61.2 & 58.2/58.3 & 59.5/59.5 \\
    \cline{1-8}
  \end{tabular}}
\label{ablation}
\end{table*}

\subsubsection{\textbf{Modality-random Disruption}} The results under different disruption rates of modality-random disruption are demonstrated in \cref{random comparison}. As the disruption rate increases from 0.2 to 1.0, the accuracy of DRF is much more stable than other methods. Under the setting of both corruption and disruption (C+D), the accuracy of previous MSA methods drops 6.72\%-9.53\% on MVSA-S, 5.00\%-6.97\% on MVSA-M, 12.78\%-18.11\% on TumEmo, indicating that the modules they devise based on prior knowledge are less effective under disruptions. For instance, MGNNS might be misled by the frequent occurrences of [MASK] tokens and bland pixels, and MVCN might suffer from inaccurate sentimental centroids caused by the disrupted modalities. Under the same setting, the accuracy of DRF only drops 4.48\% on MVSA-S, 2.52\% on MVSA-M, and 6.50\% on TumEmo. These results suggest that the sample and distribution-based recovery and quality-aware fusion facilitate the robustness of DRF to low-quality and missing modalities during both training and inference phases. 

\subsubsection{\textbf{Without Disruption}} The comparison in the regular MSA task without disruption is reported in \cref{without disruption}. DRF still achieves competitive performances against other methods. We attribute this to two reasons. Firstly, image-text pairs naturally contain modalities of different qualities. Explicitly quantifying those qualities is beneficial for the reliable fusion of modalities. Secondly, DRF learns the mapping relationships between modalities based on samples and distributions, which promotes more comprehensive information interactions between modalities.

\begin{figure*}[t]
      \centering
      \includegraphics[width=0.95\linewidth]{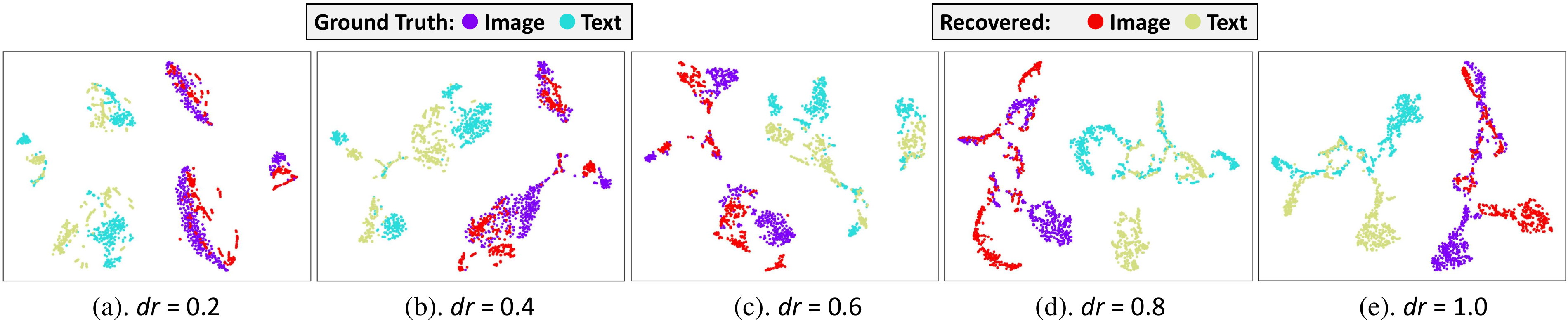}
      \vspace{-10pt}
      \caption{Visualization of image and text features on the MVSA-S test set under different disruption rates of modality-random disruption. Features are projected to 2D space by t-SNE \cite{jmlr2008tsne}.}
      \label{tsne}
      \vspace{-10pt}
\end{figure*}

\subsection{Ablation Study}
To validate the effectiveness of each key component in our method, we conduct ablation experiments under modality-fixed disruption in \cref{ablation}. From the results, we can derive the following conclusions. Firstly, both the sample-based recovery and distribution-based recovery bring performance improvements to the model, indicating that they are conducive for modality converters to learn local and global mapping relationships between modalities. Secondly, the Gaussian distribution probability and image-text pair expansion significantly facilitate the robustness of the model to low-quality modalities. It emphasizes the effectiveness of explicitly estimating modality qualities and feature fusion based on qualities. Thirdly, the image-text pair expansion also promotes the capability of the model to recover missing modalities under modality-fixed disruption. We conjecture that it introduces the sentiment prediction for recovered samples into the training process, which benefits the similar process during inference. Fourthly, the distribution constraint results in performance gains under both low-quality and missing modalities, verifying the benefits of tightening each distribution and separating different distributions. Finally, combining those components leads to the best performance, proving that they complement each other.

\subsection{Qualitative Analysis}
To intuitively present the efficacy of two recovery tasks in \cref{inter-modal recovery}, we visualize the image and text features recovered by DRF under modality-random disruption with disruption rate increases from 0.2 to 1.0. We project the samples of the MVSA-S test set into 2D space by t-SNE \cite{jmlr2008tsne} and display them in \cref{tsne}. Under low disruption rates, the recovered features closely adhere to the ground truth features. It demonstrates that DRF learns accurate mapping relationships between modalities based on the local guidance of sample-based recovery and global guidance of distribution-based recovery. As the disruption rate increases, the sample-based recovery gradually becomes unavailable, yet DRF can still recover features with distributions similar to the ground truth features. It proves the effectiveness of distribution-based recovery and emphasizes its necessity under high disruption rates.

\section{Conclusion}
In this paper, we focus on robust multimodal sentiment analysis of image-text pairs with possible low-quality and missing modalities. These issues are prevalent in real-life applications yet under-explored by previous studies in this subfield. We propose a method called DRF to handle these issues in a unified framework. It approximates the feature distributions by feature queues and leverages them to simultaneously provide global guidance for feature recovery as well as quality estimation of each modality for feature fusion. Through comprehensive experiments, we demonstrate the effectiveness and robustness of the proposed DRF.

\begin{acks}
This work is supported by the National Natural Science Foundation of China (Grant NO 62376266), and by the Key Research Program of Frontier Sciences, CAS (Grant NO ZDBS-LY-7024).
\end{acks}

\bibliographystyle{ACM-Reference-Format}
\bibliography{sample-base}

@String{Computing = "Computing" }

@String{Computer = "{IEEE} Computer" }

@String{Springer = "Springer-Verlag" }

@inproceedings{acl2019hfm,
  author       = {Yitao Cai and
                  Huiyu Cai and
                  Xiaojun Wan},
  title        = {Multi-Modal Sarcasm Detection in Twitter with Hierarchical Fusion
                  Model},
  booktitle    = {{ACL} 2019, Florence, Italy, July 28- August 2, 2019,
                  Volume 1: Long Papers},
  pages        = {2506--2515},
  publisher    = {Association for Computational Linguistics},
  year         = {2019},
  url          = {https://doi.org/10.18653/v1/p19-1239},
  doi          = {10.18653/V1/P19-1239},
  bibsource    = {dblp computer science bibliography, https://dblp.org}
}

@article{if2021survey,
  author       = {Sarah A. Abdu and
                  Ahmed H. Yousef and
                  Ashraf Salem},
  title        = {Multimodal Video Sentiment Analysis Using Deep Learning Approaches,
                  a Survey},
  journal      = {Inf. Fusion},
  volume       = {76},
  pages        = {204--226},
  year         = {2021},
  url          = {https://doi.org/10.1016/j.inffus.2021.06.003},
  doi          = {10.1016/J.INFFUS.2021.06.003},
  bibsource    = {dblp computer science bibliography, https://dblp.org}
}

@inproceedings{mm2019dialog,
  author       = {Liqiang Nie and
                  Wenjie Wang and
                  Richang Hong and
                  Meng Wang and
                  Qi Tian},
  title        = {Multimodal Dialog System: Generating Responses via Adaptive Decoders},
  booktitle    = {{MM} 2019, Nice, France, October 21-25, 2019},
  pages        = {1098--1106},
  publisher    = {{ACM}},
  year         = {2019},
  url          = {https://doi.org/10.1145/3343031.3350923},
  doi          = {10.1145/3343031.3350923},
  bibsource    = {dblp computer science bibliography, https://dblp.org}
}

@inproceedings{aaai2021hci,
  author       = {Suping Zhou and
                  Jia Jia and
                  Zhiyong Wu and
                  Zhihan Yang and
                  Yanfeng Wang and
                  Wei Chen and
                  Fanbo Meng and
                  Shuo Huang and
                  Jialie Shen and
                  Xiaochuan Wang},
  title        = {Inferring Emotion from Large-scale Internet Voice Data: {A} Semi- 
                  supervised Curriculum Augmentation based Deep Learning Approach},
  booktitle    = {{AAAI} 2021, Virtual Event, February 2-9, 2021},
  pages        = {6039--6047},
  publisher    = {{AAAI} Press},
  year         = {2021},
  url          = {https://doi.org/10.1609/aaai.v35i7.16753},
  doi          = {10.1609/AAAI.V35I7.16753},
  bibsource    = {dblp computer science bibliography, https://dblp.org}
}

@article{pami2022analysis,
  author       = {Sicheng Zhao and
                  Xingxu Yao and
                  Jufeng Yang and
                  Guoli Jia and
                  Guiguang Ding and
                  Tat{-}Seng Chua and
                  Bj{\"{o}}rn W. Schuller and
                  Kurt Keutzer},
  title        = {Affective Image Content Analysis: Two Decades Review and New Perspectives},
  journal      = {{IEEE} Trans. Pattern Anal. Mach. Intell.},
  volume       = {44},
  number       = {10},
  pages        = {6729--6751},
  year         = {2022},
  url          = {https://doi.org/10.1109/TPAMI.2021.3094362},
  doi          = {10.1109/TPAMI.2021.3094362},
  bibsource    = {dblp computer science bibliography, https://dblp.org}
}

@article{zhang2018deep,
  author       = {Lei Zhang and
                  Shuai Wang and
                  Bing Liu},
  title        = {Deep Learning for Sentiment Analysis: {A} Survey},
  journal= {Wiley Interdisciplinary Reviews: Data Mining and Knowledge Discovery},
  volume       = {8},
  number       = {4},
  year         = {2018},
  url          = {https://doi.org/10.1002/widm.1253},
  doi          = {10.1002/WIDM.1253},
}

@article{yue2019survey,
  author       = {Lin Yue and
                  Weitong Chen and
                  Xue Li and
                  Wanli Zuo and
                  Minghao Yin},
  title        = {A Survey of Sentiment Analysis in Social Media},
  journal= {Knowledge and Information Systems},
  volume       = {60},
  number       = {2},
  pages        = {617--663},
  year         = {2019},
  url          = {https://doi.org/10.1007/s10115-018-1236-4},
  doi          = {10.1007/S10115-018-1236-4},
}

@inproceedings{cikm2017multisentibank,
  author       = {Nan Xu and
                  Wenji Mao},
  title        = {MultiSentiNet: {A} Deep Semantic Network for Multimodal Sentiment
                  Analysis},
  booktitle    = {{CIKM} 2017, Singapore, November 06 - 10, 2017},
  pages        = {2399--2402},
  publisher    = {{ACM}},
  year         = {2017},
  url          = {https://doi.org/10.1145/3132847.3133142},
  doi          = {10.1145/3132847.3133142},
  bibsource    = {dblp computer science bibliography, https://dblp.org}
}

@inproceedings{isi2017hsan,
  author       = {Nan Xu},
  title        = {Analyzing Multimodal Public Sentiment Based on Hierarchical Semantic
                  Attentional Network},
  booktitle    = {{ISI} 2017, Beijing, China, July 22-24, 2017},
  pages        = {152--154},
  publisher    = {{IEEE}},
  year         = {2017},
  url          = {https://doi.org/10.1109/ISI.2017.8004895},
  doi          = {10.1109/ISI.2017.8004895},
  bibsource    = {dblp computer science bibliography, https://dblp.org}
}

@inproceedings{sigir2018comem,
  author       = {Nan Xu and
                  Wenji Mao and
                  Guandan Chen},
  title        = {A Co-Memory Network for Multimodal Sentiment Analysis},
  booktitle    = {{SIGIR} 2018, Ann Arbor, MI,
                  USA, July 08-12, 2018},
  pages        = {929--932},
  publisher    = {{ACM}},
  year         = {2018},
  url          = {https://doi.org/10.1145/3209978.3210093},
  doi          = {10.1145/3209978.3210093},
  bibsource    = {dblp computer science bibliography, https://dblp.org}
}

@article{tmm2021mvan,
  author       = {Xiaocui Yang and
                  Shi Feng and
                  Daling Wang and
                  Yifei Zhang},
  title        = {Image-Text Multimodal Emotion Classification via Multi-View Attentional
                  Network},
  journal      = {{IEEE} Trans. Multim.},
  volume       = {23},
  pages        = {4014--4026},
  year         = {2021},
  url          = {https://doi.org/10.1109/TMM.2020.3035277},
  doi          = {10.1109/TMM.2020.3035277},
  bibsource    = {dblp computer science bibliography, https://dblp.org}
}

@inproceedings{naacl2022clmlf,
  author       = {Zhen Li and
                  Bing Xu and
                  Conghui Zhu and
                  Tiejun Zhao},
  title        = {{CLMLF:} {A} Contrastive Learning and Multi-Layer Fusion Method for
                  Multimodal Sentiment Detection},
  booktitle    = {{NAACL} 2022, Seattle, WA, United States, July 10-15, 2022},
  pages        = {2282--2294},
  publisher    = {Association for Computational Linguistics},
  year         = {2022},
  url          = {https://doi.org/10.18653/v1/2022.findings-naacl.175},
  doi          = {10.18653/V1/2022.FINDINGS-NAACL.175},
  bibsource    = {dblp computer science bibliography, https://dblp.org}
}

@inproceedings{acl2021mgnns,
  author       = {Xiaocui Yang and
                  Shi Feng and
                  Yifei Zhang and
                  Daling Wang},
  title        = {Multimodal Sentiment Detection Based on Multi-channel Graph Neural
                  Networks},
  booktitle    = {{ACL/IJCNLP} 2021, (Volume 1: Long Papers), Virtual
                  Event, August 1-6, 2021},
  pages        = {328--339},
  publisher    = {Association for Computational Linguistics},
  year         = {2021},
  url          = {https://doi.org/10.18653/v1/2021.acl-long.28},
  doi          = {10.18653/V1/2021.ACL-LONG.28},
  bibsource    = {dblp computer science bibliography, https://dblp.org}
}

@inproceedings{acl2023mvcn,
  author       = {Yiwei Wei and
                  Shaozu Yuan and
                  Ruosong Yang and
                  Lei Shen and
                  Zhangmeizhi Li and
                  Longbiao Wang and
                  Meng Chen},
  title        = {Tackling Modality Heterogeneity with Multi-View Calibration Network
                  for Multimodal Sentiment Detection},
  booktitle    = {{ACL} 2023, Toronto, Canada,
                  July 9-14, 2023},
  pages        = {5240--5252},
  publisher    = {Association for Computational Linguistics},
  year         = {2023},
  url          = {https://doi.org/10.18653/v1/2023.acl-long.287},
  doi          = {10.18653/V1/2023.ACL-LONG.287},
  bibsource    = {dblp computer science bibliography, https://dblp.org}
}

@inproceedings{mm2022upmpf,
  author       = {Yang Yu and
                  Dong Zhang and
                  Shoushan Li},
  title        = {Unified Multi-modal Pre-training for Few-shot Sentiment Analysis with
                  Prompt-based Learning},
  booktitle    = {{MM} 2022, Lisboa,
                  Portugal, October 10 - 14, 2022},
  pages        = {189--198},
  publisher    = {{ACM}},
  year         = {2022},
  url          = {https://doi.org/10.1145/3503161.3548306},
  doi          = {10.1145/3503161.3548306},
  bibsource    = {dblp computer science bibliography, https://dblp.org}
}

@inproceedings{mm2023multipoint,
  author       = {Xiaocui Yang and
                  Shi Feng and
                  Daling Wang and
                  Yifei Zhang and
                  Soujanya Poria},
  title        = {Few-shot Multimodal Sentiment Analysis Based on Multimodal Probabilistic
                  Fusion Prompts},
  booktitle    = {{MM} 2023, Ottawa, ON, Canada, 29 October 2023- 3 November 2023},
  pages        = {6045--6053},
  publisher    = {{ACM}},
  year         = {2023},
  url          = {https://doi.org/10.1145/3581783.3612181},
  doi          = {10.1145/3581783.3612181},
  bibsource    = {dblp computer science bibliography, https://dblp.org}
}

@inproceedings{nips2023incomplete,
  author       = {Yuanzhi Wang and
                  Yong Li and
                  Zhen Cui},
  title        = {Incomplete Multimodality-Diffused Emotion Recognition},
  booktitle    = {NeurIPS 2023, New Orleans,
                  LA, USA, December 10 - 16, 2023},
  year         = {2023},
  url          = {http://papers.nips.cc/paper\_files/paper/2023/hash/372cb7805eaccb2b7eed641271a30eec-Abstract-Conference.html},
  bibsource    = {dblp computer science bibliography, https://dblp.org}
}

@inproceedings{ijcai2018semi-incomplete,
  author       = {Yang Yang and
                  De{-}Chuan Zhan and
                  Xiang{-}Rong Sheng and
                  Yuan Jiang},
  title        = {Semi-Supervised Multi-Modal Learning with Incomplete Modalities},
  booktitle    = {{IJCAI} 2018, July 13-19, 2018, Stockholm,
                  Sweden},
  pages        = {2998--3004},
  publisher    = {ijcai.org},
  year         = {2018},
  url          = {https://doi.org/10.24963/ijcai.2018/416},
  doi          = {10.24963/IJCAI.2018/416},
  bibsource    = {dblp computer science bibliography, https://dblp.org}
}

@inproceedings{nips2019cpmnet,
  author       = {Changqing Zhang and
                  Zongbo Han and
                  Yajie Cui and
                  Huazhu Fu and
                  Joey Tianyi Zhou and
                  Qinghua Hu},
  title        = {CPM-Nets: Cross Partial Multi-View Networks},
  booktitle    = {NeurIPS 2019, December
                  8-14, 2019, Vancouver, BC, Canada},
  pages        = {557--567},
  year         = {2019},
  url          = {https://proceedings.neurips.cc/paper/2019/hash/11b9842e0a271ff252c1903e7132cd68-Abstract.html},
  bibsource    = {dblp computer science bibliography, https://dblp.org}
}

@inproceedings{aaai2019cycle-translate,
  author       = {Hai Pham and
                  Paul Pu Liang and
                  Thomas Manzini and
                  Louis{-}Philippe Morency and
                  Barnab{\'{a}}s P{\'{o}}czos},
  title        = {Found in Translation: Learning Robust Joint Representations by Cyclic
                  Translations between Modalities},
  booktitle    = {{AAAI} 2019, Honolulu, Hawaii,
                  USA, January 27 - February 1, 2019},
  pages        = {6892--6899},
  publisher    = {{AAAI} Press},
  year         = {2019},
  url          = {https://doi.org/10.1609/aaai.v33i01.33016892},
  doi          = {10.1609/AAAI.V33I01.33016892},
  bibsource    = {dblp computer science bibliography, https://dblp.org}
}

@article{pami2023gcnet,
  author       = {Zheng Lian and
                  Lan Chen and
                  Licai Sun and
                  Bin Liu and
                  Jianhua Tao},
  title        = {GCNet: Graph Completion Network for Incomplete Multimodal Learning
                  in Conversation},
  journal      = {{IEEE} Trans. Pattern Anal. Mach. Intell.},
  volume       = {45},
  number       = {7},
  pages        = {8419--8432},
  year         = {2023},
  url          = {https://doi.org/10.1109/TPAMI.2023.3234553},
  doi          = {10.1109/TPAMI.2023.3234553},
  bibsource    = {dblp computer science bibliography, https://dblp.org}
}

@inproceedings{iclr2021tmvc,
  author       = {Zongbo Han and
                  Changqing Zhang and
                  Huazhu Fu and
                  Joey Tianyi Zhou},
  title        = {Trusted Multi-View Classification},
  booktitle    = {{ICLR} 2021,
                  Virtual Event, Austria, May 3-7, 2021},
  publisher    = {OpenReview.net},
  year         = {2021},
  url          = {https://openreview.net/forum?id=OOsR8BzCnl5},
  bibsource    = {dblp computer science bibliography, https://dblp.org}
}

@article{pami2022tmvc,
  author       = {Zongbo Han and
                  Changqing Zhang and
                  Huazhu Fu and
                  Joey Tianyi Zhou},
  title        = {Trusted Multi-View Classification With Dynamic Evidential Fusion},
  journal      = {{IEEE} Trans. Pattern Anal. Mach. Intell.},
  volume       = {45},
  number       = {2},
  pages        = {2551--2566},
  year         = {2023},
  url          = {https://doi.org/10.1109/TPAMI.2022.3171983},
  doi          = {10.1109/TPAMI.2022.3171983},
  bibsource    = {dblp computer science bibliography, https://dblp.org}
}

@incollection{1967dst,
  author       = {Arthur P. Dempster},
  title        = {Upper and Lower Probabilities Induced by a Multivalued Mapping},
  booktitle    = {Classic Works of the Dempster-Shafer Theory of Belief Functions},
  series       = {Studies in Fuzziness and Soft Computing},
  volume       = {219},
  pages        = {57--72},
  publisher    = {Springer},
  year         = {2008},
  url          = {https://doi.org/10.1007/978-3-540-44792-4\_3},
  doi          = {10.1007/978-3-540-44792-4\_3},
  bibsource    = {dblp computer science bibliography, https://dblp.org}
}

@inproceedings{acl2021mmin,
  author       = {Jinming Zhao and
                  Ruichen Li and
                  Qin Jin},
  title        = {Missing Modality Imagination Network for Emotion Recognition with
                  Uncertain Missing Modalities},
  booktitle    = {{ACL/IJCNLP} 2021, (Volume 1: Long Papers), Virtual
                  Event, August 1-6, 2021},
  pages        = {2608--2618},
  publisher    = {Association for Computational Linguistics},
  year         = {2021},
  url          = {https://doi.org/10.18653/v1/2021.acl-long.203},
  doi          = {10.18653/V1/2021.ACL-LONG.203},
  bibsource    = {dblp computer science bibliography, https://dblp.org}
}

@inproceedings{mmm2016mvsa,
  author       = {Teng Niu and
                  Shiai Zhu and
                  Lei Pang and
                  Abdulmotaleb El{-}Saddik},
  title        = {Sentiment Analysis on Multi-View Social Data},
  booktitle    = {{MMM} 2016, Miami,
                  FL, USA, January 4-6, 2016, Proceedings, Part {II}},
  series       = {Lecture Notes in Computer Science},
  volume       = {9517},
  pages        = {15--27},
  publisher    = {Springer},
  year         = {2016},
  url          = {https://doi.org/10.1007/978-3-319-27674-8\_2},
  doi          = {10.1007/978-3-319-27674-8\_2},
  bibsource    = {dblp computer science bibliography, https://dblp.org}
}

@article{tsap2012speech1,
  author       = {Nelson Morgan},
  title        = {Deep and Wide: Multiple Layers in Automatic Speech Recognition},
  journal      = {{IEEE} Trans. Speech Audio Process.},
  volume       = {20},
  number       = {1},
  pages        = {7--13},
  year         = {2012},
  url          = {https://doi.org/10.1109/TASL.2011.2116010},
  doi          = {10.1109/TASL.2011.2116010},
  bibsource    = {dblp computer science bibliography, https://dblp.org}
}

@inproceedings{isca2017speech2,
  author       = {Duc Le and
                  Zakaria Aldeneh and
                  Emily Mower Provost},
  title        = {Discretized Continuous Speech Emotion Recognition with Multi-Task
                  Deep Recurrent Neural Network},
  booktitle    = {Interspeech 2017, Stockholm, Sweden, August 20-24, 2017},
  pages        = {1108--1112},
  publisher    = {{ISCA}},
  year         = {2017},
  url          = {https://doi.org/10.21437/Interspeech.2017-94},
  doi          = {10.21437/INTERSPEECH.2017-94},
  bibsource    = {dblp computer science bibliography, https://dblp.org}
}

@article{ftir2007text1,
  author       = {Bo Pang and
                  Lillian Lee},
  title        = {Opinion Mining and Sentiment Analysis},
  journal      = {Found. Trends Inf. Retr.},
  volume       = {2},
  number       = {1-2},
  pages        = {1--135},
  year         = {2007},
  url          = {https://doi.org/10.1561/1500000011},
  doi          = {10.1561/1500000011},
  bibsource    = {dblp computer science bibliography, https://dblp.org}
}

@article{cl2011text2,
  author       = {Maite Taboada and
                  Julian Brooke and
                  Milan Tofiloski and
                  Kimberly D. Voll and
                  Manfred Stede},
  title        = {Lexicon-Based Methods for Sentiment Analysis},
  journal      = {Comput. Linguistics},
  volume       = {37},
  number       = {2},
  pages        = {267--307},
  year         = {2011},
  url          = {https://doi.org/10.1162/COLI\_a\_00049},
  doi          = {10.1162/COLI\_A\_00049},
  bibsource    = {dblp computer science bibliography, https://dblp.org}
}

@inproceedings{icip2008image1,
  author       = {Victoria Yanulevskaya and
                  Jan C. van Gemert and
                  Katharina Roth and
                  Ann{-}Katrin Herbold and
                  Nicu Sebe and
                  Jan{-}Mark Geusebroek},
  title        = {Emotional Valence Categorization using Holistic Image Features},
  booktitle    = {{ICIP} 2008, October 12-15, 2008, San Diego, California, {USA}},
  pages        = {101--104},
  publisher    = {{IEEE}},
  year         = {2008},
  url          = {https://doi.org/10.1109/ICIP.2008.4711701},
  doi          = {10.1109/ICIP.2008.4711701},
  bibsource    = {dblp computer science bibliography, https://dblp.org}
}

@inproceedings{aaai2015image2,
  author       = {Quanzeng You and
                  Jiebo Luo and
                  Hailin Jin and
                  Jianchao Yang},
  title        = {Robust Image Sentiment Analysis Using Progressively Trained and Domain
                  Transferred Deep Networks},
  booktitle    = {{AAAI} 2015, January 25-30, 2015, Austin, Texas, {USA}},
  pages        = {381--388},
  publisher    = {{AAAI} Press},
  year         = {2015},
  url          = {https://doi.org/10.1609/aaai.v29i1.9179},
  doi          = {10.1609/AAAI.V29I1.9179},
  bibsource    = {dblp computer science bibliography, https://dblp.org}
}

@inproceedings{nips2017transformer,
  author       = {Ashish Vaswani and
                  Noam Shazeer and
                  Niki Parmar and
                  Jakob Uszkoreit and
                  Llion Jones and
                  Aidan N. Gomez and
                  Lukasz Kaiser and
                  Illia Polosukhin},
  title        = {Attention is All you Need},
  booktitle    = {NeurIPS 2017, December 4-9, 2017,
                  Long Beach, CA, {USA}},
  pages        = {5998--6008},
  year         = {2017},
  url          = {https://proceedings.neurips.cc/paper/2017/hash/3f5ee243547dee91fbd053c1c4a845aa-Abstract.html},
  bibsource    = {dblp computer science bibliography, https://dblp.org}
}

@inproceedings{ijcai2019tombert,
  author       = {Jianfei Yu and
                  Jing Jiang},
  title        = {Adapting {BERT} for Target-Oriented Multimodal Sentiment Classification},
  booktitle    = {{IJCAI} 2019, Macao, China, August 10-16,
                  2019},
  pages        = {5408--5414},
  publisher    = {ijcai.org},
  year         = {2019},
  url          = {https://doi.org/10.24963/ijcai.2019/751},
  doi          = {10.24963/IJCAI.2019/751},
  bibsource    = {dblp computer science bibliography, https://dblp.org}
}

@inproceedings{mm2021captrbert,
  author       = {Zaid Khan and
                  Yun Fu},
  title        = {Exploiting {BERT} for Multimodal Target Sentiment Classification through
                  Input Space Translation},
  booktitle    = {{MM} 2021, Virtual Event, China, October
                  20 - 24, 2021},
  pages        = {3034--3042},
  publisher    = {{ACM}},
  year         = {2021},
  url          = {https://doi.org/10.1145/3474085.3475692},
  doi          = {10.1145/3474085.3475692},
  bibsource    = {dblp computer science bibliography, https://dblp.org}
}

@inproceedings{acl2022vlpmabsa,
  author       = {Yan Ling and
                  Jianfei Yu and
                  Rui Xia},
  title        = {Vision-Language Pre-Training for Multimodal Aspect-Based Sentiment
                  Analysis},
  booktitle    = {{ACL} 2022, Dublin, Ireland,
                  May 22-27, 2022},
  pages        = {2149--2159},
  publisher    = {Association for Computational Linguistics},
  year         = {2022},
  url          = {https://doi.org/10.18653/v1/2022.acl-long.152},
  doi          = {10.18653/V1/2022.ACL-LONG.152},
  bibsource    = {dblp computer science bibliography, https://dblp.org}
}

@inproceedings{emnlp2022fite,
  author       = {Hao Yang and
                  Yanyan Zhao and
                  Bing Qin},
  title        = {Face-Sensitive Image-to-Emotional-Text Cross-modal Translation for
                  Multimodal Aspect-based Sentiment Analysis},
  booktitle    = {{EMNLP} 2022, Abu Dhabi, United Arab Emirates,
                  December 7-11, 2022},
  pages        = {3324--3335},
  publisher    = {Association for Computational Linguistics},
  year         = {2022},
  url          = {https://doi.org/10.18653/v1/2022.emnlp-main.219},
  doi          = {10.18653/V1/2022.EMNLP-MAIN.219},
  bibsource    = {dblp computer science bibliography, https://dblp.org}
}

@book{shafer1976dst,
  title={A Mathematical Theory of Evidence},
  author={Shafer, Glenn},
  volume={42},
  year={1976},
  publisher={Princeton university press}
}

@inproceedings{nips1990bnn,
  author       = {John S. Denker and
                  Yann LeCun},
  title        = {Transforming Neural-Net Output Levels to Probability Distributions},
  booktitle    = {NeurIPS 1990,
                  Denver, Colorado, USA, November 26-29, 1990},
  pages        = {853--859},
  publisher    = {Morgan Kaufmann},
  year         = {1990},
  url          = {http://papers.nips.cc/paper/419-transforming-neural-net-output-levels-to-probability-distributions},
  bibsource    = {dblp computer science bibliography, https://dblp.org}
}

@inproceedings{icml2016dropout,
  author       = {Yarin Gal and
                  Zoubin Ghahramani},
  title        = {Dropout as a Bayesian Approximation: Representing Model Uncertainty
                  in Deep Learning},
  booktitle    = {{ICML} 2016, New York City, NY, USA, June 19-24, 2016},
  series       = {{JMLR} Workshop and Conference Proceedings},
  volume       = {48},
  pages        = {1050--1059},
  publisher    = {JMLR.org},
  year         = {2016},
  url          = {http://proceedings.mlr.press/v48/gal16.html},
  bibsource    = {dblp computer science bibliography, https://dblp.org}
}

@inproceedings{nips2017ensemble,
  author       = {Balaji Lakshminarayanan and
                  Alexander Pritzel and
                  Charles Blundell},
  title        = {Simple and Scalable Predictive Uncertainty Estimation using Deep Ensembles},
  booktitle    = {NeurIPS 2017, December 4-9, 2017,
                  Long Beach, CA, {USA}},
  pages        = {6402--6413},
  year         = {2017},
  url          = {https://proceedings.neurips.cc/paper/2017/hash/9ef2ed4b7fd2c810847ffa5fa85bce38-Abstract.html},
  bibsource    = {dblp computer science bibliography, https://dblp.org}
}

@inproceedings{iclr2021ensemble2,
  author       = {Marton Havasi and
                  Rodolphe Jenatton and
                  Stanislav Fort and
                  Jeremiah Zhe Liu and
                  Jasper Snoek and
                  Balaji Lakshminarayanan and
                  Andrew Mingbo Dai and
                  Dustin Tran},
  title        = {Training independent subnetworks for robust prediction},
  booktitle    = {{ICLR} 2021,
                  Virtual Event, Austria, May 3-7, 2021},
  publisher    = {OpenReview.net},
  year         = {2021},
  url          = {https://openreview.net/forum?id=OGg9XnKxFAH},
  bibsource    = {dblp computer science bibliography, https://dblp.org}
}

@inproceedings{nips2021monig,
  author       = {Huan Ma and
                  Zongbo Han and
                  Changqing Zhang and
                  Huazhu Fu and
                  Joey Tianyi Zhou and
                  Qinghua Hu},
  title        = {Trustworthy Multimodal Regression with Mixture of Normal-inverse Gamma
                  Distributions},
  booktitle    = {NeurIPS 2021, December
                  6-14, 2021, virtual},
  pages        = {6881--6893},
  year         = {2021},
  url          = {https://proceedings.neurips.cc/paper/2021/hash/371bce7dc83817b7893bcdeed13799b5-Abstract.html},
  bibsource    = {dblp computer science bibliography, https://dblp.org}
}

@inproceedings{nips2020energy1,
  author       = {Weitang Liu and
                  Xiaoyun Wang and
                  John D. Owens and
                  Yixuan Li},
  title        = {Energy-based Out-of-distribution Detection},
  booktitle    = {NeurIPS 2020, December
                  6-12, 2020, virtual},
  year         = {2020},
  url          = {https://proceedings.neurips.cc/paper/2020/hash/f5496252609c43eb8a3d147ab9b9c006-Abstract.html},
  bibsource    = {dblp computer science bibliography, https://dblp.org}
}

@inproceedings{icml2023qmf,
  author       = {Qingyang Zhang and
                  Haitao Wu and
                  Changqing Zhang and
                  Qinghua Hu and
                  Huazhu Fu and
                  Joey Tianyi Zhou and
                  Xi Peng},
  title        = {Provable Dynamic Fusion for Low-Quality Multimodal Data},
  booktitle    = {{ICML} 2023, 23-29 July
                  2023, Honolulu, Hawaii, {USA}},
  series       = {Proceedings of Machine Learning Research},
  volume       = {202},
  pages        = {41753--41769},
  publisher    = {{PMLR}},
  year         = {2023},
  url          = {https://proceedings.mlr.press/v202/zhang23ar.html},
  bibsource    = {dblp computer science bibliography, https://dblp.org}
}

@article{pami2022fixpad1,
  author       = {Changqing Zhang and
                  Yajie Cui and
                  Zongbo Han and
                  Joey Tianyi Zhou and
                  Huazhu Fu and
                  Qinghua Hu},
  title        = {Deep Partial Multi-View Learning},
  journal      = {{IEEE} Trans. Pattern Anal. Mach. Intell.},
  volume       = {44},
  number       = {5},
  pages        = {2402--2415},
  year         = {2022},
  url          = {https://doi.org/10.1109/TPAMI.2020.3037734},
  doi          = {10.1109/TPAMI.2020.3037734},
  bibsource    = {dblp computer science bibliography, https://dblp.org}
}

@inproceedings{kdd2020fixpad2,
  author       = {Jiayi Chen and
                  Aidong Zhang},
  title        = {{HGMF:} Heterogeneous Graph-based Fusion for Multimodal Data with
                  Incompleteness},
  booktitle    = {{KDD} 2020, Virtual Event, CA, USA, August 23-27, 2020},
  pages        = {1295--1305},
  publisher    = {{ACM}},
  year         = {2020},
  url          = {https://doi.org/10.1145/3394486.3403182},
  doi          = {10.1145/3394486.3403182},
  bibsource    = {dblp computer science bibliography, https://dblp.org}
}

@article{siam2010low-rank1,
  author       = {Jian{-}Feng Cai and
                  Emmanuel J. Cand{\`{e}}s and
                  Zuowei Shen},
  title        = {A Singular Value Thresholding Algorithm for Matrix Completion},
  journal      = {{SIAM} J. Optim.},
  volume       = {20},
  number       = {4},
  pages        = {1956--1982},
  year         = {2010},
  url          = {https://doi.org/10.1137/080738970},
  doi          = {10.1137/080738970},
  bibsource    = {dblp computer science bibliography, https://dblp.org}
}

@article{jmlr2010low-rank2,
  author       = {Rahul Mazumder and
                  Trevor Hastie and
                  Robert Tibshirani},
  title        = {Spectral Regularization Algorithms for Learning Large Incomplete Matrices},
  journal      = {J. Mach. Learn. Res.},
  volume       = {11},
  pages        = {2287--2322},
  year         = {2010},
  url          = {https://dl.acm.org/doi/10.5555/1756006.1859931},
  doi          = {10.5555/1756006.1859931},
  bibsource    = {dblp computer science bibliography, https://dblp.org}
}

@inproceedings{icml2008autoencoder,
  author       = {Pascal Vincent and
                  Hugo Larochelle and
                  Yoshua Bengio and
                  Pierre{-}Antoine Manzagol},
  title        = {Extracting and composing robust features with denoising autoencoders},
  booktitle    = {{ICML} 2008, Helsinki, Finland, June 5-9, 2008},
  series       = {{ACM} International Conference Proceeding Series},
  volume       = {307},
  pages        = {1096--1103},
  publisher    = {{ACM}},
  year         = {2008},
  url          = {https://doi.org/10.1145/1390156.1390294},
  doi          = {10.1145/1390156.1390294},
  bibsource    = {dblp computer science bibliography, https://dblp.org}
}

@inproceedings{cvpr2020moco,
  author       = {Kaiming He and
                  Haoqi Fan and
                  Yuxin Wu and
                  Saining Xie and
                  Ross B. Girshick},
  title        = {Momentum Contrast for Unsupervised Visual Representation Learning},
  booktitle    = {{CVPR} 2020, Seattle, WA, USA, June 13-19, 2020},
  pages        = {9726--9735},
  publisher    = {Computer Vision Foundation / {IEEE}},
  year         = {2020},
  url          = {https://doi.org/10.1109/CVPR42600.2020.00975},
  doi          = {10.1109/CVPR42600.2020.00975},
  bibsource    = {dblp computer science bibliography, https://dblp.org}
}

@inproceedings{cvpr2018id,
  author       = {Zhirong Wu and
                  Yuanjun Xiong and
                  Stella X. Yu and
                  Dahua Lin},
  title        = {Unsupervised Feature Learning via Non-Parametric Instance Discrimination},
  booktitle    = {{CVPR} 2018, Salt Lake City, UT, USA, June 18-22, 2018},
  pages        = {3733--3742},
  publisher    = {Computer Vision Foundation / {IEEE} Computer Society},
  year         = {2018},
  url          = {http://openaccess.thecvf.com/content\_cvpr\_2018/html/Wu\_Unsupervised\_Feature\_Learning\_CVPR\_2018\_paper.html},
  doi          = {10.1109/CVPR.2018.00393},
  bibsource    = {dblp computer science bibliography, https://dblp.org}
}

@inproceedings{iclr2021vit,
  author       = {Alexey Dosovitskiy and
                  Lucas Beyer and
                  Alexander Kolesnikov and
                  Dirk Weissenborn and
                  Xiaohua Zhai and
                  Thomas Unterthiner and
                  Mostafa Dehghani and
                  Matthias Minderer and
                  Georg Heigold and
                  Sylvain Gelly and
                  Jakob Uszkoreit and
                  Neil Houlsby},
  title        = {An Image is Worth 16x16 Words: Transformers for Image Recognition
                  at Scale},
  booktitle    = {{ICLR} 2021,
                  Virtual Event, Austria, May 3-7, 2021},
  publisher    = {OpenReview.net},
  year         = {2021},
  url          = {https://openreview.net/forum?id=YicbFdNTTy},
  bibsource    = {dblp computer science bibliography, https://dblp.org}
}

@inproceedings{naacl2019bert,
  author       = {Jacob Devlin and
                  Ming{-}Wei Chang and
                  Kenton Lee and
                  Kristina Toutanova},
  title        = {{BERT:} Pre-training of Deep Bidirectional Transformers for Language
                  Understanding},
  booktitle    = {{NAACL-HLT} 2019, Minneapolis, MN, USA, June 2-7, 2019, Volume 1 (Long
                  and Short Papers)},
  pages        = {4171--4186},
  publisher    = {Association for Computational Linguistics},
  year         = {2019},
  url          = {https://doi.org/10.18653/v1/n19-1423},
  doi          = {10.18653/V1/N19-1423},
  bibsource    = {dblp computer science bibliography, https://dblp.org}
}

@inproceedings{acl2019mult,
  author       = {Yao{-}Hung Hubert Tsai and
                  Shaojie Bai and
                  Paul Pu Liang and
                  J. Zico Kolter and
                  Louis{-}Philippe Morency and
                  Ruslan Salakhutdinov},
  title        = {Multimodal Transformer for Unaligned Multimodal Language Sequences},
  booktitle    = {{ACL} 2019, Florence, Italy, July 28- August 2, 2019,
                  Volume 1: Long Papers},
  pages        = {6558--6569},
  publisher    = {Association for Computational Linguistics},
  year         = {2019},
  url          = {https://doi.org/10.18653/v1/p19-1656},
  doi          = {10.18653/V1/P19-1656},
  bibsource    = {dblp computer science bibliography, https://dblp.org}
}

@article{jmlr2008tsne,
  title={Visualizing Data using t-SNE.},
  author={Van der Maaten, Laurens and Hinton, Geoffrey},
  journal={Journal of Machine Learning Research},
  volume={9},
  number={11},
  year={2008}
}

@inproceedings{mm2021tbfr,
  author       = {Ziqi Yuan and
                  Wei Li and
                  Hua Xu and
                  Wenmeng Yu},
  title        = {Transformer-based Feature Reconstruction Network for Robust Multimodal
                  Sentiment Analysis},
  booktitle    = {{MM} 2021, Virtual Event, China, October
                  20 - 24, 2021},
  pages        = {4400--4407},
  publisher    = {{ACM}},
  year         = {2021},
  url          = {https://doi.org/10.1145/3474085.3475585},
  doi          = {10.1145/3474085.3475585},
}

@article{taffc2024emt,
  author       = {Licai Sun and
                  Zheng Lian and
                  Bin Liu and
                  Jianhua Tao},
  title        = {Efficient Multimodal Transformer With Dual-Level Feature Restoration
                  for Robust Multimodal Sentiment Analysis},
  journal      = {{IEEE} Trans. Affect. Comput.},
  volume       = {15},
  number       = {1},
  pages        = {309--325},
  year         = {2024},
  url          = {https://doi.org/10.1109/TAFFC.2023.3274829},
  doi          = {10.1109/TAFFC.2023.3274829},
}

@inproceedings{emnlp2022mmalign,
  author       = {Wei Han and
                  Hui Chen and
                  Min{-}Yen Kan and
                  Soujanya Poria},
  title        = {MM-Align: Learning Optimal Transport-based Alignment Dynamics for
                  Fast and Accurate Inference on Missing Modality Sequences},
  booktitle    = {{EMNLP} 2022, Abu Dhabi, United Arab Emirates,
                  December 7-11, 2022},
  pages        = {10498--10511},
  publisher    = {Association for Computational Linguistics},
  year         = {2022},
  url          = {https://doi.org/10.18653/v1/2022.emnlp-main.717},
  doi          = {10.18653/V1/2022.EMNLP-MAIN.717},
}

@article{tmm2024nibat,
  author       = {Ziqi Yuan and
                  Yihe Liu and
                  Hua Xu and
                  Kai Gao},
  title        = {Noise Imitation Based Adversarial Training for Robust Multimodal Sentiment
                  Analysis},
  journal      = {{IEEE} Trans. Multim.},
  volume       = {26},
  pages        = {529--539},
  year         = {2024},
  url          = {https://doi.org/10.1109/TMM.2023.3267882},
  doi          = {10.1109/TMM.2023.3267882},
}

@inproceedings{fcy2022icme,
  author       = {Chengyang Fang and
                  Gangyan Zeng and
                  Yu Zhou and
                  Daiqing Wu and
                  Can Ma and
                  Dayong Hu and
                  Weiping Wang},
  title        = {Towards Escaping from Language Bias and {OCR} Error: Semantics-Centered
                  Text Visual Question Answering},
  booktitle    = {{ICME} 2022, Taipei, Taiwan, July 18-22, 2022},
  pages        = {1--6},
  publisher    = {{IEEE}},
  year         = {2022},
  url          = {https://doi.org/10.1109/ICME52920.2022.9859603},
  doi          = {10.1109/ICME52920.2022.9859603},
}

@article{wdq2024arxiv,
  author       = {Daiqing Wu and
                  Dongbao Yang and
                  Huawen Shen and
                  Can Ma and
                  Yu Zhou},
  title        = {Resolving Sentiment Discrepancy for Multimodal Sentiment Detection via Semantics Completion and Decomposition},
  journal      = {CoRR},
  volume       = {abs/2407.07026},
  year         = {2024},
  url          = {https://doi.org/10.48550/arXiv.2407.07026},
  doi          = {10.48550/ARXIV.2407.07026},
  eprinttype    = {arXiv},
  eprint       = {2407.07026},
}

@inproceedings{shw2023ijcai,
  author       = {Huawen Shen and
                  Xiang Gao and
                  Jin Wei and
                  Liang Qiao and
                  Yu Zhou and
                  Qiang Li and
                  Zhanzhan Cheng},
  title        = {Divide Rows and Conquer Cells: Towards Structure Recognition for Large
                  Tables},
  booktitle    = {{IJCAI} 2023, 19th-25th August 2023, Macao,
                  SAR, China},
  pages        = {1369--1377},
  publisher    = {ijcai.org},
  year         = {2023},
  url          = {https://doi.org/10.24963/ijcai.2023/152},
  doi          = {10.24963/IJCAI.2023/152},
}


\end{document}